\documentclass{bmvc2k}

\title{Mixstyle-Entropy: Domain Generalization with Causal Intervention and Perturbation\footnote{$^{*}$First two authors contributed equally.}\footnote{$^{\dag}$Corresponding author.}}

\addauthor{Luyao Tang$^{*}$}{http://lytang63.github.io}{1}
\addauthor{Yuxuan Yuan$^{*}$}
{yuanyuxuan0908@stu.xmu.edu.cn}{1}
\addauthor{Chaoqi Chen}{cqchen1994@gmail.com}{2}
\addauthor{Xinghao Ding}{dxh@xmu.edu.cn}{1}
\addauthor{Yue Huang$^{\dag}$}{yhuang2010@xmu.edu.cn}{1}

\addinstitution{
 School of Informatics\\
 Xiamen University\\
 Xiamen, China
}
\addinstitution{
 Department of Computer Science\\
 The University of Hong Kong\\
 Hong Kong, China
}

\runninghead{Luyao TANG ET AL}{DG via Causal Intervention and Perturbation}

\usepackage{amsthm,amsmath,amssymb}
\usepackage{multirow}
\usepackage{MnSymbol}
\usepackage{mathrsfs}
\usepackage{caption}
\usepackage{booktabs} 
\usepackage{wrapfig}
\usepackage{graphicx}
\usepackage{subcaption}
\usepackage{amsmath}
\usepackage{multicol}

\begin{document}

\maketitle

\begin{abstract}
Despite the considerable advancements achieved by deep neural networks, their performance tends to degenerate when the test environment diverges from the training ones. Domain generalization~(DG) solves this issue by learning representations independent of domain-related information, thus facilitating extrapolation to unseen environments. Existing approaches typically focus on formulating tailored training objectives to extract shared features from the source data. However, the disjointed training and testing procedures may compromise robustness, particularly in the face of unforeseen variations during deployment. In this paper, we propose a novel and holistic framework based on causality, named \texttt{InPer}, designed to enhance model generalization by incorporating causal intervention during training and causal perturbation during testing. Specifically, during the training phase, we employ entropy-based causal intervention (EnIn) to refine the selection of causal variables. To identify samples with anti-interference causal variables from the target domain, we propose a novel metric, homeostatic score, through causal perturbation (HoPer) to construct a prototype classifier in test time. Experimental results across multiple cross-domain tasks confirm the efficacy of \texttt{InPer}. Code is available at: \href{https://github.com/lytang63/Inper}{github.com/lytang63/Inper}.

\end{abstract}

\section{Introduction}
\label{sec:intro}

Deep neural networks (DNNs) have achieved remarkable success in various computer vision applications, including image classification and object detection tasks. However, Nonetheless, their training relies on the assumption that the training and testing datasets are identically and independently distributed (IID)~\cite{ben2010theory}. In real-world scenarios, this assumption is often violated, leading to a marked degradation in the performance of models trained on the source domain~\cite{hendrycks2019benchmarking, recht2019imagenet}.

Domain generalization (DG) aims to enhance the generalization capacity of deep neural networks (DNNs) towards out-of-distribution (OOD) data. The challenge of OOD scenarios \cite{Koh2020WILDSAB} has been tackled from diverse perspectives encompassing optimization, model architecture, and data manipulation. Optimization-wise, diverse training strategies have sought to cultivate domain-independent feature representations. These strategies include explicit feature alignment~\cite{Ghifary2015DomainGF,Jin2020StyleNA,Motiian2017UnifiedDS, chen2022compound} and adversarial learning~\cite{10.5555/3045118.3045244,Gong2018DLOWDF}. Concerning model design, enhancing generalization often involves meticulous crafting of DNN structures \cite{DBLP:conf/aaai/Liu0LZZ22,DBLP:journals/corr/abs-1807-09441,DBLP:journals/corr/abs-1907-04275} or employing ensembles of multiple expert models \cite{Mancini2018BestSF,Zhou2020DomainAE}. On the data front, techniques like data augmentation~\cite{Nazari2020DomainGU, Tremblay2018TrainingDN, Yue2019DomainRA} and generation are leveraged to enrich the diversity~\cite{Shankar2018GeneralizingAD, Zhou2020DeepDI, chen2022relation} of training samples. Additionally, approaches based on causal learning~\cite{Christiansen2020ACF,lv2022causality, chen2022mix} and meta-learning~\cite{Wei2021ToAlignTA, chen2022compound} have also been explored. 

However, mainstream DG strategies often overlook two pivotal limitations. Firstly, while many methods prioritize domain invariance~\cite{li2018domain, dou2019domain, zhang2022exact, li2021consistent}, they often disregard spurious correlations within features. For example, enforcing domain invariance can introduce spurious correlations. Mixstyle~\cite{zhou2021domain} randomly selects instances and applies linear interpolation to their statistical features, overlooking spurious correlations between semantic content and image style. Secondly, expanding the feature space blindly can lead to intricate decision boundaries. DSU~\cite{li2022uncertainty} utilizes multivariate Gaussian distributions to construct virtual instances. However, such an approach can compromise existing class embeddings, resulting in reduced inter-class distances and rugged decision boundaries, as shown in Figure~\ref{fig:tsne}. 

From the perspective of causal learning, the aforementioned approaches share a common objective of extracting domain-invariant variables. A more insightful approach would be: the process of extracting domain-invariant features should involve modeling data generation, extracting causal relationships from observable variables~\cite{mahajan2021domain}, and isolating causal variables used for classification.

Based on the discussion above, we aim to differentiate genuine domain-related variables and intervene in the forward process. 
To this end, we propose \texttt{InPer}, a straightforward yet efficacious DG approach. \texttt{InPer} consists of two key components: Intervention and Perturbation. 
During training, we extract domain-related statistical features using theoretically derived feature entropy. We perform causal interventions on samples in the embedding space to sever the association between domain-related information and causal variables. This disentanglement enables the isolation of causal variables containing semantic information crucial for precise prediction. 
Moreover, we integrate causal learning into the model deployment phase through Causal Perturbation, which adjusts embeddings towards the centroid of their respective classes. Additionally, we propose a novel Homeostatic Score, to assess the interference resilience of causal variable branches in a structured causal graph. Finally, we construct a prototype classifier suitable for the testing domain using samples with stable causal relationships and progressively fit the target distribution.


\section{Related Work}
\label{sec:related_work}

\subsection{Causality and Domain Generalization}

Causality is often used to establish connections between causal relationships and model generalization~\cite{peters2017elements,christiansen2021causal}. Various techniques such as invariant causal mechanisms~\cite{wang2022contrastive, heinze2021conditional, subbaswamy2019preventing} and restoring causal features~\cite{liu2021heterogeneous, rojas2018invariant} have been proposed to boost OOD generalization. Earlier work~\cite{gong2016domain} ventured causal reasoning into domain adaptation, and others aimed to establish causal links between class labels and samples~\cite{magliacane2018domain, heinze2021conditional, peters2016causal, li2022domain}. MatchDG~\cite{mahajan2021domain} suggested a domain generalization invariance condition via causal flow between labels. However, many causality and domain generalization solutions rely heavily on restrictive assumptions about causal graphs or structural equations, such as CIRL~\cite{lv2022causality} employs dimensional representations to mimic causal factors, depending on only a highly generic causal structure model without restrictive assumptions.

\subsection{Online Parameter Updating}

Online parameter updating, a technique typically improving model performance in unfamiliar testing domains, involves testing distribution alignment operations. The chief approaches include Test-time training (TTT)~\cite{liu2021ttt++, sun2020test} and Test-time adaptation (TTA)~\cite{boudiaf2022parameter, wang2020tent, niu2022efficient, iwasawa2021test}. TTT enhances the model through self-supervised testing data tasks like rotation classification~\cite{komodakis2018unsupervised}. TTA adjusts the model during testing without altering training, with Tent~\cite{wang2020tent} minimizing entropy for parameter updates, and SHOT~\cite{liang2020we} maximizing mutual information. Some studies~\cite{niu2022efficient, wang2022continual, chen2023coda} explored TTA and continual learning amalgamation.

\begin{figure*}[t]
	\begin{center}
	\includegraphics[width=1.00\textwidth]{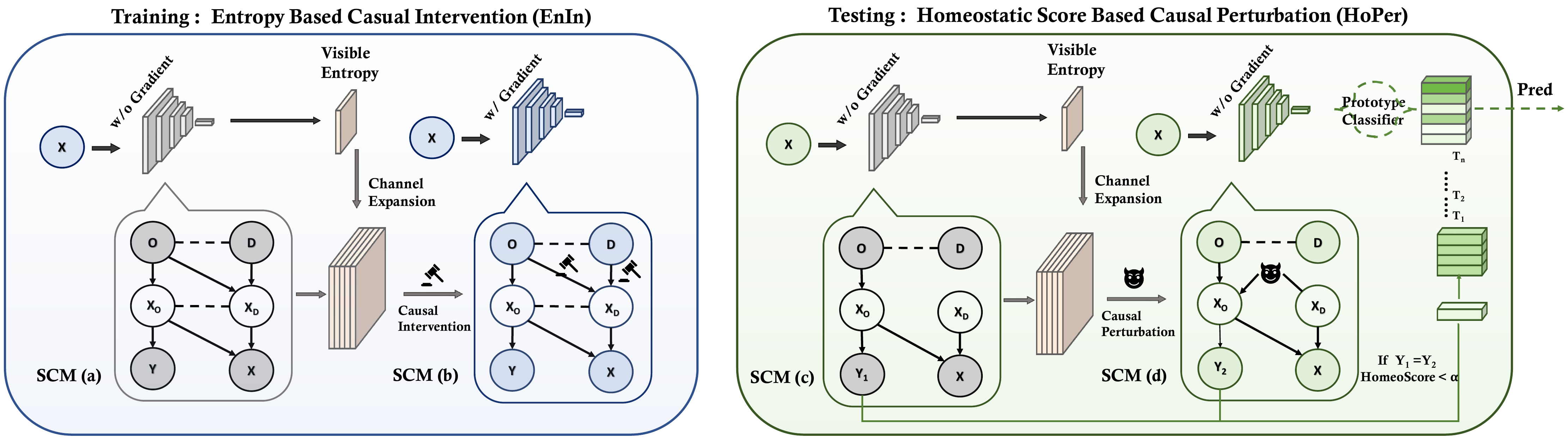}
 \vspace{0.0cm}
		\caption{The pipeline of the proposed \texttt{InPer}.
  \textbf{Training:} Extract visible entropy and perform causal interventions on samples to sever the spurious connections between class and domain. 
         \textbf{Testing:} Causal perturbation is applied to the testing samples, and through the HomeoScore, a generalized classifier is built to adapt to the target domain.}
		\label{fig:overall}
	\end{center}
\end{figure*}

\section{Methodology}
\label{sec:methodology}

\subsection{Theoretical Insight}

In this section, we explain the theoretical foundations of our work. Similar to the causal relationship constructed by MatchDG~\cite{mahajan2021domain}, we present a Structural Causal Model (SCM)~\cite{10.1145/3501714.3501755} for DG task in Figure~\ref{fig:overall} SCM (a). For intuition, let's consider an example inspired by Mixstyle~\cite{zhou2021domain}. Each domain (D) displays a unique 'style' for images from different domains while the various objects (O) exhibit a semantic characteristic 'shape'. 
$X_O$ are the high-level causal features of $O$ which induce the label $Y$. Note that another high-level proxy feature $X_{D}$ in SCM is determined by the style of the given image. At the same time, this style information comes from both $D$ and $O$.
Due to the common parent node $O$, $X_O$ and $X_D$ are correlated, and $D$ will inevitably influence the prediction of $X_O$. However, we expect that the class $Y$ only relies on the casual feature $X_O$ of the object category. The SCM (a) can be expressed as:
\begin{equation}
    \label{eq:scm}
    \begin{aligned}
        \text{SCM (a)}:\ \ 
        X_O&:=f_{X_O}\left(O, U_{X_O}\right) \ \ \ \ \ \ \ X_D:=f_{X_D}\left(O, D, U_{X_D}\right) \\ 
        X&:=f_X\left(X_O, X_D, U_X\right) \ \ \ \ 
        Y:=f_Y\left(X_O, U_Y\right). \\
    \end{aligned}
\end{equation}
The variables $\{X_O, X_D, X, Y\}$ are the endogenous variables~\cite{kruglanski1975endogenous}, and they have parent nodes in causal graphs. Additionally, the implicit noise variable $U$, which includes all relevant background conditions for the deterministic variable, is omitted from the causal graphs. Note that the structural equation is a generalized expression of the variable conditional probability distribution, the truncated factorization~\cite{pearl2018book} and backdoor adjustment~\cite{pearl2018book} techniques remain applicable.

In the task of DG, a model is trained on the source training set $\mathcal{D}_s = \{\mathcal{D}_1, \ldots, \mathcal{D}_S\} (S \ge 2)$, and evaluated on the unseen target set 
The n-th source domain is denoted as $\mathcal{D}_n=\left\{\left(\mathbf{x}_i, y_i, d_i\right)\right\}_{i=1}^{N_n}$, where $\mathbf{x}_i \in \mathbb{R}^D$ represents the training data, $y_i =\{1,2, \ldots, K\}$ is the class label, $d_i=\{1,2,\ldots, S\}$ denotes the domain label. Source and target domains follow different data distributions in the joint space $\mathcal{X} \times \mathcal{Y}$. Therefore, the optimization function (ERM)~\cite{vapnik1991principles} can be represented as follows:
\begin{equation}
\label{eq:erm}
    \arg \min \mathbb{E}_{\left(\mathbf{x}_i, y_i, d_i\right) \in \mathcal{D}_s}[\ell\left(f\left(\mathbf{x}_i\right), y_i\right)],
\end{equation}
where $\ell(\cdot, \cdot)$ quantifies the inconsistency between between predictions and labels.

To enhance the performance of domain-agnostic feature representations $g(\mathrm{x})$
we employ an entropy-based formulation as done by distribution-matching methods~\cite{akuzawa2020adversarial}. 
We follow the methodology in MatchDG~\cite{mahajan2021domain} and aim to learn a representation that is independent of domain given class label. It can be interpreted as maximizing the entropy of domain given class label, i.e, $g^*(x)=\arg \max H(d|y, g(\mathrm{x}))$, the optimal $g^*(x)$ satisfies $H\left(d|y, g^*(x)\right)=H(d|y)$.

Unlike MatchDG follows the assumption that $x_o \upmodels d|y$, we consider a more realistic scenario whereby $x_o \nupmodels d|y$, i.e., $P(x_o|y)$ varies across the domains. A simple example is that although sharing the same label, the dogs depicted in the sketch exhibit significant differences in their morphological characteristics compared to those in the painting.
In this case, considering the correlation between $X_O$ and $X_D$ in SCM (a), we obtain the following:
\begin{equation}
    \label{eq:entropy1}
    H(g(\mathrm{x}), d|y) - H(d|y) \leqslant H(x_o, x_d, d|y) - H(d|y).
\end{equation}
If $x_o \upmodels x_d$, the above equation can be simplified to:
\begin{equation}
    \begin{aligned}
    \label{eq:entropy2}
        H(g(\mathrm{x}), d|y) - H(d|y) &= H(x_o, x_d, d|y) - H(d|y) = H(x_o, d|y) - H(d|y) = KL(x_o, d|y).
     \end{aligned}
\end{equation}

Turning our attention to node $X_D$ under the aforementioned DG setup, $x_d$ changes across domains, i.e., $x_d \nupmodels y|d$. Similar to the above analysis, we can also perform a causal analysis on $x_d$ and obtain the formula for entropy. For simplicity, we denote the entropy feature representation as $H(g(\mathrm{x}))$. The relationship between $H(g(\mathrm{x}))$ and variables $x_o$ and $x_d$ can be expressed by the following formula:

\begin{equation}
    \label{eq:entropy3}
    \begin{aligned}
        H(g(\mathrm{x})) \propto KL(x_o, d|y)  \ \ \ \   
        H(g(\mathrm{x}))\propto KL(x_d, y|d).
    \end{aligned}
\end{equation}

In general, our goal is to minimize prediction error using the most 'pure' $x_o$. We achieve this by performing a causal intervention, using the do-operation $do(\cdot)$~\cite{pearl2018book} on $x_d$. We aim to remove domain-relevant information from the feature representation similar to CIRL~\cite{lv2022causality} and MixStyle~\cite{zhou2021domain}. Ideally, we can eliminate the arrow pointing to $X_D$, clip the causal relationship between $O$ and $X_D$, and remove the association between $X_O$ and $X_D$ in SCM (a).
The d-separation and perfect map assumption~\cite{pearl2009causality} are employed to obtain the category-related semantic factors for prediction. Two critical aspects need to be considered, i.e., the specific form of $do(\cdot)$ and the assumption of $x_d$. 
In CIRL, the intervention is performed on $x_d$ by mixing the amplitude information from other domains after Fourier transformation. However, due to the spurious correlation between $O$ and $D$, the ${x_o'}$ of the intervened sample will be introduced into the ${x_o''}$ of nother samples.
Therefore, careful selection of causal intervention variables is crucial for improving the model's generalization performance. Based on the theoretical analysis above, we detail our approach in Section \ref{EnIn} and \ref{HoPer}.

\subsection{Training: Entropy Based Casual Intervention}
\label{EnIn}

Taking CNN as an example, the model $f(\mathrm{x})=\textbf{W}^{\top} g(\mathrm{x})$ can be decomposed into a feature extractor $g(\cdot)$ and a classifier. $g(\cdot)$ consists of multiple blocks with a stacking depth of $m$, namely, $g(\cdot)=[g_1(\cdot), g_2(\cdot), \ldots, g_m(\cdot)]$. As the input passes through every block, the dimension of the feature space is expanded progressively until the local feature vector $\textbf{v}_{\textbf{i}}$ is obtained for classification. Here, $\textbf{v}_{\textbf{i}} \in \mathbb{R}^c$ with indices $i=\left\{1,2, \ldots, h w\right\}$. The final prediction scores $\textbf{F}$ for $K$ classes are computed before applying the softmax function in the following way:
\begin{equation}
\label{eq:predscore}
    \textbf{F}=\textbf{W}^{\top} \frac{1}{h w} \sum_i \textbf{v}_i=\frac{1}{h w} \sum_i \textbf{W}^{\top} \textbf{v}_i.
\end{equation}
We aim to display the entropy of different local feature vectors $\textbf{v}_{\textbf{i}}$ using varying weights. We denote $\textbf{W}^{\top} \textbf{v}_{\textbf{i}}$ as $\hat{\textbf{F}}_i \in \mathbb{R}^K$. It represents the local class score vector of location $i$ in the feature map. 
The semantic information contained in each $\textbf{v}_i$ is extracted before the pooling operation. $\hat{\textbf{p}}_i = \text{softmax}(\hat{\textbf{F}}_i)$ is the local class probability at location $i$. To compute the Shannon entropy~\cite{shannon2001mathematical} of $\hat{\textbf{p}}_i$, we use $\label{eq:entropyofpred}
H\left(\hat{\textbf{p}}_i\right)=-\sum_{k=1}^K \hat{\textbf{p}}_i(k) \log \hat{\textbf{p}}_i(k)$.
Then we use a simple normalization function $\operatorname{Normalize}(x):=\frac{x-\min x}{\max x-\min x},$  to map the elements to the range [0,1]. This normalization process provides numerical information about the features, which is utilized to generate the feature entropy mask: $\mathcal{M}=\text {Normalize}\left(H\left(\hat{\textbf{p}}_i\right)\right)$.

Then, channel duplication and spatial interpolation are applied in $\mathcal{M}$ to achieve scaling without introducing extra parameters. Taking inspiration from normalization technique~\cite{zhou2021domain, nuriel2021permuted},
the statistics of features from instances to be observable and manipulable $x_d$ to achieve causal intervention. During mini-batch training, features $g(\mathrm{x})$ are randomly sampled from the training data. Subsequently, $\tilde{g}(\mathrm{x})$ is obtained by $\operatorname{shuffle}$, the specific formula of $\operatorname{shuffle}$, $\sigma(\cdot)$, and $\mu(\cdot)$ are shown in the supplementary material. We discussed in Eq.~\ref{eq:entropy2} that $x_d$ variables in features with the maximum entropy and domain-related information is the most relevant. Hence, we must filter the statistical features from $\tilde{g}(\mathrm{x})$ to a significant extent to ensure that $x_d$ comes from its domain. The simplest solution is to crop the featuremap and its corresponding entropy into patches of equivalent sizes. Then choose the region with the maximum entropy $\tilde{\mathcal{M}}_{crop}^{max}$ and the minimum entropy $\tilde{\mathcal{M}}_{crop}^{min}$, calculate mixed feature statistics using the following:
\begin{equation}
    \begin{aligned}
        \tilde{\gamma}_{mix} & =\lambda \sigma(g(\mathrm{x}))+(1-\lambda) \sigma(\tilde{\mathcal{M}}_{crop}^{max} \odot \tilde{g}(\mathrm{x}) ) \ \ 
        \tilde{\beta}_{mix} =\lambda \mu(g(\mathrm{x}))+(1-\lambda) \mu(\tilde{\mathcal{M}}_{crop}^{max} \odot \tilde{g}(\mathrm{x}) ) \\
        \gamma_{mix} & =\lambda \sigma(g(\mathrm{x}))+(1-\lambda) \sigma(\mathcal{M}_{crop}^{min} \odot g(\mathrm{x}))  \ \ 
        \beta_{mix} =\lambda \mu(g(\mathrm{x}))+(1-\lambda) \mu(\mathcal{M}_{crop}^{min} \odot g(\mathrm{x}) ),
    \end{aligned}
\end{equation}
where $\lambda \sim \operatorname{Beta}(\alpha, \alpha)$. The motivation for constructing $\mathcal{M}_{crop}^{max}$ comes from the second half of Eq.~\ref{eq:entropy3} to achieve a better causal intervention plan and minimize the inclusion of class-related variables from $\tilde{g}(\mathrm{x})$ in the intervention sample, thereby preventing blurring of the classification boundary in the feature space. The first half of Eq.~\ref{eq:entropy3} shows the relationship between entropy, the causal variable $x_o$, and domain-related information. The optimization goal for the current intervention sample should also consider the relationship $x_o \upmodels d|y$. In addition, we can strengthen the semantic features of the samples themselves by $\mathcal{M}_{crop}^{min}$. Therefore, the mini-batch EnIn is formulated as follows:
\begin{equation}
    \label{eq:fred}
    \operatorname{EnIn}(g(\mathrm{x}))=
    \begin{cases}
    \mathcal{M} \odot g(\mathrm{x}) +
    \tilde{\gamma}_{\operatorname{mix}} \frac{g(\mathrm{x})-\mu(g(\mathrm{x}))}{\sigma(g(\mathrm{x}))}+\tilde{\beta}_{\text {mix}} \\
    \gamma_{\operatorname{mix}} \frac{g(\mathrm{x})-\mu(g(\mathrm{x}))}{\sigma(g(\mathrm{x}))}+\beta_{\text {mix}}.
    \end{cases}
\end{equation}

 For simplification, during the forward pass of mini-batch training, $\operatorname{EnIn}$ is executed with a probability of 0.5 without gradients, and not executed during the test.

\subsection{Testing: Homeostatic Score Based Causal Perturbation}
\label{HoPer}
Most DG algorithms prioritize the training stage by focusing on extracting domain-independent information from multiple source domains. Nevertheless, it is also crucial to deliberate how to model the features of unlabeled data rationally during the testing phase, to counter the issue of domain shift. Recent efforts~\cite{wang2020tent,liang2020we} utilize backpropagation to train the model on the target domain or construct a prototype classifier and adjust it during testing~\cite{iwasawa2021test, chen2023activate}.
However, they neglect two aspects: 1) the samples used for constructing category prototypes may inadequately represent the current category characteristics; 2) the samples affecting the decision boundary are still disregarded. 



Using the causal graph constructed above, we build a prototype classifier using causal stable samples. Specifically, we modify the $x_d$ attributes of test samples, introducing more class-related $x_o$ information. This generates a feature representation closer to the category centroid, facilitating a denser cluster of category prototypes. Instead of only filtering high-entropy samples~\cite{iwasawa2021test}, which can limit generalization performance, we introduce the Homeostatic Score (HomeoScore) for sample selection. By comparing the probability difference between original and perturbed test samples, we exclude perturbation-sensitive samples and optimize the decision boundary.

During testing, we use a memory bank $\mathbb{B}=\left\{\left(g'(\mathrm{x}), p{'}\right)\right\}$ to retain sample embedding features and predicted logits. We sample and extract feature representations $g(\mathrm{x_t})$ along with their class probabilities $p_t$ and pseudo-label $y_t$ at time $t$. As these representations are based on the source domain feature extractor, guaranteeing their generalizability is challenging. Therefore, we apply a feature transformation similar to EnIn via Eq.~\ref{eq:entropy3}, where $\gamma_{mix} =\lambda \sigma(g(\mathrm{x_t}))+(1-\lambda) \sigma(\mathcal{M}_{crop}^{min} \odot g(\mathrm{x_t}))$ and $ \beta_{mix} =\lambda \mu(g(\mathrm{x_t}))+(1-\lambda) \mu(\mathcal{M}_{crop}^{min} \odot g(\mathrm{x_t}) )$, to obtain the target domain's new representation $g{'}(\mathrm{x_t})$ using Causal Perturbation (CP). CP boosts class-related information in $g(\mathrm{x_t})$, generating perturbed $p_t{'}$ and $y_t{'}$, and resulting in a tighter cluster of prototypes in the memory bank. Outliers are then filtered using the HomeoScore, defined as the distance between the original and perturbed sample representations:

\begin{equation}
    \operatorname{HomeoScore} = \left(\sum_{j=1}^k\left|p_t^j - {p_t^j}{'}\right|^2\right)^{\frac{1}{2}}.
\end{equation}
Then the prototype of class $k$ could be formulated as :
\begin{equation}
    \mathbb{B}_t^k= \begin{cases}\mathbb{B}_{t-1}^k \cup\left\{\frac{g{'}(\mathrm{x_{t-1}})}{\left\|{g{'}(\mathrm{x_{t-1}})}\right\|}\right\} & ,\text { if } y'_{(t-1)}=y^k \ and \  \operatorname{HomeoScore} < \alpha \\ 
    \mathbb{B}_{t-1}^k & ,\text { else } \end{cases}
\end{equation}
where $\alpha$ is a fixed threshold, which is set to $0.2$. Figure~\ref{fig:bar-and-line} illustrates that $\operatorname{HomeoScore}$ can better distinguish erroneous pseudo-labels. We refrain from using samples with a larger $\operatorname{HomeoScore}$ for updating the memory bank as their pseudo-labels can damage the decision-making of the prototype classifier. The entropy threshold $\beta$ controls the resource usage and operational efficiency of the memory bank using $\mathbb{B}_t^k=\left\{g{'}(\mathrm{x}) \mid g{'}(\mathrm{x}) \in \mathbb{B}_t^k, H(p{'}) \leq \beta \right\}$. Finally, we define the prototype-based classification output as the softmax over the feature similarities to prototypes for class $k$ :
\begin{equation}
    y_j^k=\frac{\exp \left(sim\left(g_j(\mathrm{x_t}), c_k\right)\right)}{\sum_{k^{\prime}=1}^{|Y|} \exp \left(sim\left(g_j(\mathrm{x_t}), c_{k^{\prime}}\right)\right)},
\end{equation}
where $c^k=\frac{1}{\left|\mathbb{B}^k\right|} \sum_{\boldsymbol{z} \in \mathbb{B}_t^k} \boldsymbol{z}$
, $sim\left(\cdot, \cdot \right)$ denotes cosine similarity.

\section{Experiments}
\label{experiments}

To validate the generalization performance of our proposed method. We conduct experiments on image classification, semantic segmentation, and instance retrieval, specifically where the training and testing sets are from different domains. More experiment results and implementation details are presented in the supplementary material. All results are generated through five rounds of experiments with different random seeds.

\subsection{Generalization on Multi-Source Domain Classification}

\textbf{Datasets and Implementation Details.}
We conduct classification experiments based on Dassl~\cite{zhou2021domain} and verify the generalization performance on four standard DG datasets PACS~\cite{li2017deeper} and Office-home~\cite{venkateswara2017deep}. PACS consists of four different domains, each domain containing 7 object categories. Office-Home spans four domains across 65 categories. 

\begin{table}[t]
\begin{center}
\resizebox{\textwidth}{!}{
    \begin{tabular}{l|cccc|c|cccc|c}
        \hline
         & \multicolumn{5}{c|}{PACS} & \multicolumn{5}{c}{Office-Home}\\
        \hline
        Method & Art & Cartoon & Photo & Sketch & Avg.(\%) & Art & Clipart & Product & Real & Avg.(\%) \\
        \hline
        \text {Baseline}  & 74.3 & 76.7 & \textbf{96.4} & 68.7 & 79.0 & 58.8 & 48.3 & 74.2 & 76.2 & 64.4 \\
        \text {Mixup~\cite{zhang2018mixup}} & 76.8 & 74.9 & 95.8 & 66.6 & 78.5 & 58.2 & 49.3 & 74.7 & 76.1 & 64.6 \\
        \text {RSC~\cite{huang2020self}} & 78.9 & 76.9 & 94.1 & 76.8 & 81.7 & 58.4 & 47.9 & 71.6 & 74.5 & 63.1 \\
        \text {L2A-OT~\cite{zhou2020learning}} & 83.3 & 78.2 & 96.2 & 76.3 & 82.8 & 60.6 & 50.1 & 74.8 & 77.0 & 65.6 \\
        \text {MixStyle~\cite{zhou2021domain}} & 82.3 & 79.0 & \underline{96.3} & 73.8 & 82.8 & 58.7 & 53.4 & 74.2 & 75.9 & 65.5 \\
        \text {DSU~\cite{li2022uncertainty}} & 83.6 & 79.6 & 95.8 & 77.6 & 84.1 & 60.2 & 54.8 & 74.1 & 75.1 & 66.1 \\
        \text {CIRL~\cite{lv2022causality}} & 86.1 & 80.6 & 95.9 & 82.7 & 86.3 & 61.5 & 55.3 & 75.1 & 76.6 & 67.1 \\
        \hline
        \text {\texttt{InPer}\ (Ours)} & \textbf{88.5}$\pm$0.4 & \textbf{84.2}$\pm$0.3 & 95.3$\pm$0.2 & \textbf{85.8}$\pm$0.5 & \textbf{88.5}$\pm$0.3 & \textbf{67.0}$\pm$0.4 & \textbf{62.3}$\pm$0.4 & \textbf{75.5}$\pm$0.4 & \textbf{77.3}$\pm$0.2 & \textbf{70.5}$\pm$0.3 \\
        \textcolor{red}{-} \text {HomeoScore} & \underline{88.0}$\pm$0.5 & \underline{83.6}$\pm$0.4 & 95.1$\pm$0.2 & \underline{84.8}$\pm$0.3 & \underline{87.9}$\pm$0.3 & \underline{66.7}$\pm$0.5 & \underline{61.8}$\pm$0.6 & \underline{74.7}$\pm$0.4 & \underline{77.0}$\pm$0.2 & \underline{70.1}$\pm$0.3 \\
        \textcolor{red}{-} \text {HoPer} & 86.8$\pm$0.3 & 82.5$\pm$0.3 & 94.9$\pm$0.6 & 82.3$\pm$0.2 & 86.7$\pm$0.4 & 65.7$\pm$0.5 & 60.3$\pm$0.3 & 73.7$\pm$0.5 & 76.6$\pm$0.4 & 69.1$\pm$0.4 \\
        \hline
    \end{tabular}
}
\end{center}
\caption{Leave-one-domain-out multi-domain classification on ResNet-18.}
\label{tab:r18}
\end{table}

\begin{table}[t]
\begin{center}
\resizebox{\textwidth}{!}
{
    \begin{tabular}{l|cccc|c|cccc|c}
        \hline
         & \multicolumn{5}{c|}{PACS} & \multicolumn{5}{c}{Office-Home}\\
        \hline
        Method & Art & Cartoon & Photo & Sketch & Avg.($\%$) & Art & Clipart & Product & Real & Avg.($\%$) \\
        \hline
        Baseline & 86.2 & 78.7 & \underline{97.6} & 70.6 & 83.2 & 61.3 & 52.4 & 75.8 & 76.6 & 66.5 \\
        RSC~\cite{huang2020self} & 87.8 & 82.1 & \textbf{97.9} & 83.3 & 87.9 & 50.7 & 51.4 & 74.8 & 75.1 & 65.5 \\
        SelfReg~\cite{kim2021selfreg} & 87.9 & 79.4 & 96.8 & 78.3 & 85.6 & 63.6 & 53.1 & 76.9 & 78.1 & 67.9 \\
        SagNet~\cite{nam2021reducing} & 81.1 & 75.4 & 95.7 & 77.2 & 82.3 & 63.4 & 54.8 & 75.8 & 78.3 & 68.1 \\
        I$^{2}$\text{-ADR}~\cite{meng2022attention} & 88.5 & 83.2 & 95.2 & 85.8 & 88.2 & 70.3 & 55.1 & 80.7 & 79.2 & 71.4 \\
        \hline
        \text {\texttt{InPer}\ (Ours)} & $\textbf{91.9}\pm0.4$ & $\textbf{87.6}\pm0.6$ &$ 96.9\pm0.3$ &$\textbf{87.5}\pm0.7$ & $\textbf{91.0}\pm0.5$ & $\textbf{74.4}\pm0.4$& $\textbf{66.8}\pm0.5$ & $\textbf{80.6}\pm0.4$ & $\textbf{81.6}\pm0.4$ & $\textbf{75.9}\pm0.4$ \\
        \textcolor{red}{-} \text {HomeoScore} &$\underline{91.3}\pm0.5$ &$\underline{86.8}\pm0.5$ & $96.7\pm0.2$&$\underline{87.1}\pm0.4$& $\underline{90.5}\pm0.3$  & $\underline{73.9}\pm0.5$ & $\underline{66.2}\pm0.4$ & $\underline{79.7}\pm0.3$ & $\underline{80.9}\pm0.5$ &$\underline{75.1}\pm0.4$ \\
        \textcolor{red}{-} \text {HoPer} & $90.9\pm0.5$& $86.2\pm0.5$ & $96.5\pm0.3$&$86.7\pm0.4$ & $90.1\pm0.4$  & $73.1\pm0.6$& $65.3\pm0.4 $& $78.9\pm0.3$ & $80.1\pm0.3 $&$74.4\pm0.4$ \\
        \hline
    \end{tabular}
}
\end{center}
\caption{Leave-one-domain-out multi-domain classification on ResNet-50.}
\label{tab:r50}
\end{table}


\noindent 
\textbf{Results on PACS and Office-Home} based on ReNet-18 and ResNet-50 are reported in Table~\ref{tab:r18} and Table~\ref{tab:r50}, respectively. It can be observed that $\texttt{InPer}$ reaches the highest average accuracy among all the compared methods on both backbones even without HoPer. For
 PACS, compared with CIRL~\cite{lv2022causality}, $\texttt{InPer}$ outperforms it by a large margin of 2.2 \% and on ResNet-18. Our approach is also superior to the ensemble method I\textsuperscript{2}-ADR~\cite{meng2022attention} on the parameter-rich ResNet-50, achieving a higher accuracy of 2.8 \% improvement. For Office-Home, our approach exhibits a 3.4 \% improvement over CIRL on ResNet-18 and achieves a 4.5 \% improvement over the previous SOTA~\cite{meng2022attention} on ResNet-50.



\noindent 
\textbf{Highlights.} $\texttt{InPer}$ does not require any additional training parameters. Compared to previous methods that expand the embedding space, such as MixStyle~\cite{zhou2021domain} and DSU~\cite{lv2022causality}, our method consistently surpasses them in terms of average accuracy by 5.3 \% and 4.4 \%, respectively. Even after removing the HoPer module, our approach still achieves SOTA performance, surpassing them by 3.7 \% and 2.8\%. Notably, $\texttt{InPer}$ can be seamlessly integrated into any existing DG framework.

\subsection{Generalization on Semantic Segmentation}

\textbf{Datasets and Implementation Details.}
Semantic segmentation impacts autonomous driving, yet dynamic driving scenes can hamper it. We assess $\texttt{InPer}$ generalizability on the gaming-based GTA5~\cite{richter2016playing} image dataset and the Cityscapes~\cite{cordts2016cityscapes} dataset featuring German urban street scenes. Our implementation aligns with prior work DSU~\cite{lv2022causality}.

\begin{figure*}[t]
	\begin{center}
	\includegraphics[width=0.95\textwidth]{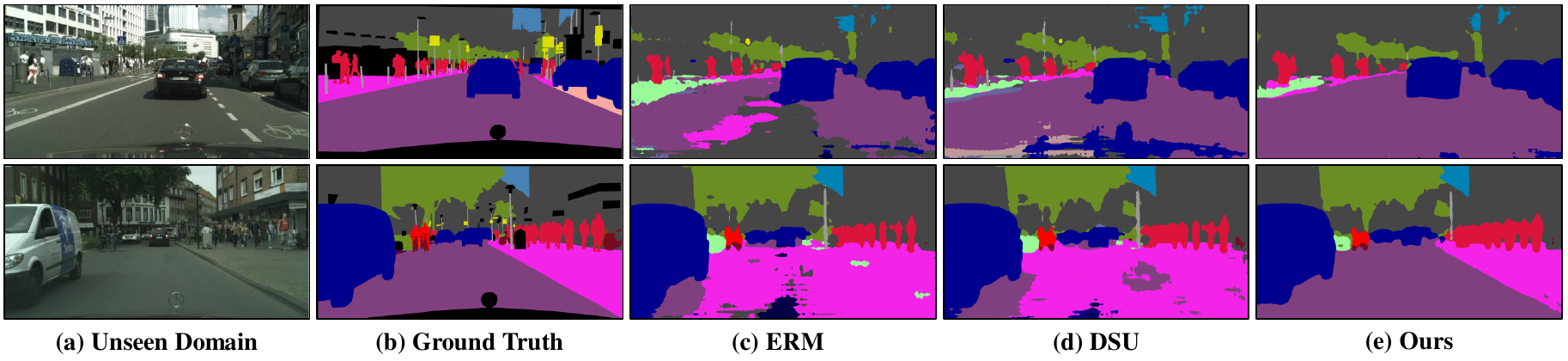}
 \vspace{0.5cm}
\caption{The visualization on unseen domain Cityscapes with the model trained on GTA5.}
	\label{fig:seg}
	\end{center}
\end{figure*}

 \begin{wraptable}{r!}{0.40\textwidth}
\resizebox{\linewidth}{!}{
 \begin{tabular}{c|ccc}
        \hline
        Method & mIoU(\%) & mAcc(\%) \\
         \hline 
        ERM &36.0&49.5\\
        pAdaIN~\cite{nuriel2021permuted}&38.1&51.1\\
        MixStyle~\cite{zhou2021domain}&38.7&51.2\\
        DSU~\cite{li2022uncertainty}&\underline{40.7}&$\mathbf{53.8}$\\
        \hline
        EnIn (Ours) &$\mathbf{42.6}$ &\underline{53.7}\\
        \hline
    \end{tabular}
    }
     \vspace{0.2cm}
    \caption{Results of semantic segmentation from GAT5 to Cityscapes.\label{tab:seg}}
        \vspace{-0.2cm}
        
\end{wraptable}


\noindent 
\textbf{Results on Semantic Segmentation} are shown in Table~\ref{tab:seg}. It emphasizes that appropriate feature space expansion is vital in pixel-level classification. Overstepping other categories' embedding spaces could lead to incorrect predictions, illustrated in Figure~\ref{fig:seg}. In sunny conditions, vehicle shadows can lead to car misclassifications. Only EnIn accurately segments the car region by breaking this false association. The failure to differentiate sidewalks from road surfaces in pixel predictions shows the danger of indiscreet feature space interpolation, particularly in autonomous driving cases. Only our method correctly segments such scenes, providing clear differentiation.


\begin{figure*}[t]
	\begin{center}
	\includegraphics[width=0.95\textwidth]{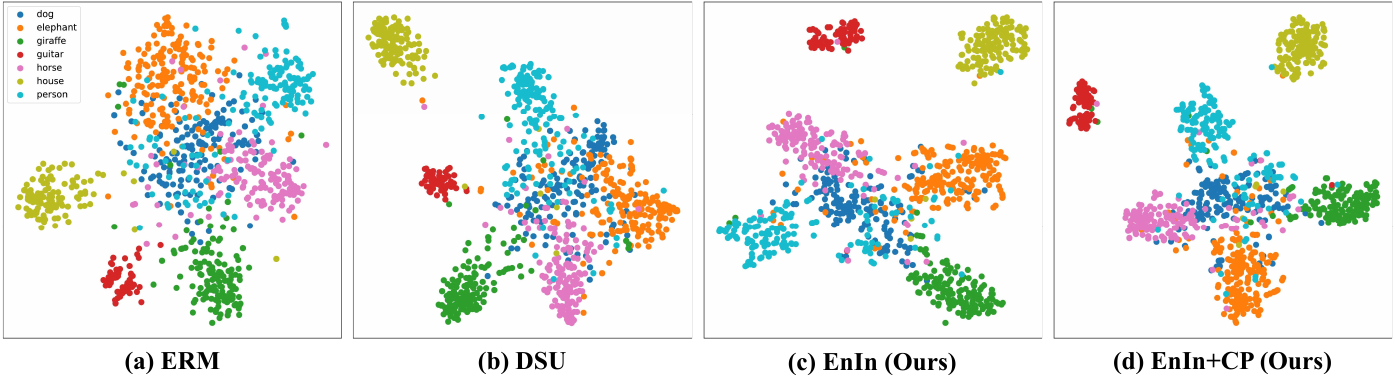}
 \vspace{0.5cm}
\caption{The t-SNE visualization of extracted deep features using different methods on PACS dataset (target domain: cartoon). The different colors stands for different classes.}
	\label{fig:tsne}
	\end{center}
\end{figure*}

\begin{figure*}[t]
	\begin{center}
	\includegraphics[width=1.00\textwidth]{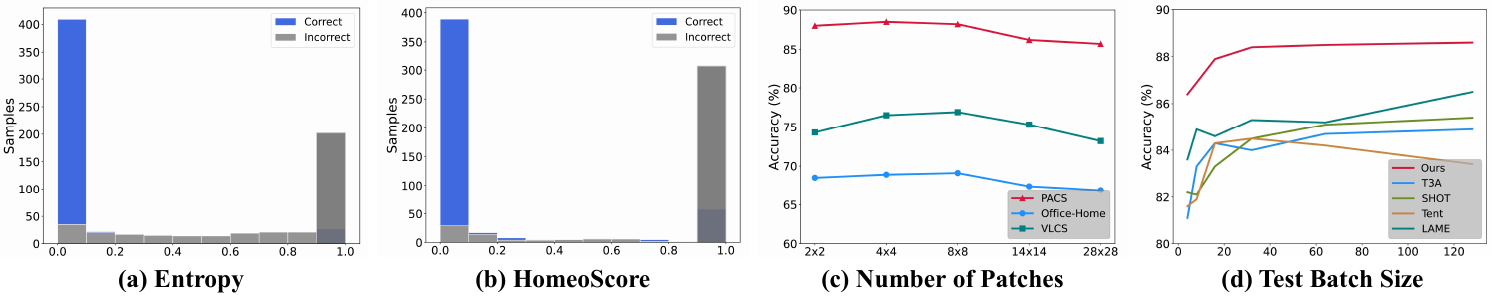}
 \vspace{0.0cm}
\caption{(a) and (b) are pseudo-label selection capability between entropy and HomenScore. (c) is effects of patch sizes. (d) is the influence of test batch size in different schemes.}
        \vspace{-0.5cm} 
	\label{fig:bar-and-line}
	\end{center}
\end{figure*}


\section{Analysis and Ablations}


\textbf{Ablation Study.} $\texttt{InPer}$, designed as a plug-and-play module with minimal hyperparameters, hinges on two components: EnIn and HoPer. HomeoScore filtering in HoPer improves average accuracy by 0.6\% through providing superior samples, as seen in Tables~\ref{tab:r18} and~\ref{tab:r50}. HomeoScore outperforms traditional Entropy in pseudo-label filtering (Figure~\ref{fig:bar-and-line}). Incorrectly predicted test domain samples possess higher HomeoScore, hence, filtering these optimizes the prototype classifier decision boundary. Despite HoPer exclusion during testing, $\texttt{InPer}$ surpasses most previous works.

\noindent 
\textbf{Effects of Causal Perturbation.}
During the testing phase, we apply causal perturbation (CP) to the samples to further inject class-related information. To better demonstrate the sample distribution in the feature space, we employ t-SNE~\cite{van2008visualizing} to visualize the embeddings of the test domain samples, as shown in Figure~\ref{fig:tsne}. DSU~\cite{lv2022causality} using multidimensional Gaussian distributions to simulate the distributions, leads to decreased inter-class distances and blurry decision boundaries. In contrast, EnIn implicitly measures the inter-cluster and intra-cluster distances through the feature entropy representation, resulting in an expansion of inter-cluster distances. After undergoing CP, the intra-cluster distances decrease, leading to more compact clusters and clear classification boundaries for the model.

\noindent 
\textbf{Effects of Patch size and Test Batch Size}
The selection of $\mathcal{M}$ is based on patching the feature maps. It was found that a ratio of $\frac{1}{4}$ or $\frac{1}{8}$ resulted in better performance on all three datasets in Figure~\ref{fig:bar-and-line}. Further reducing the ratio led to a significant decline in model performance due to insufficient statistical features to represent $x_o$. Additionally, we compared $\texttt{InPer}$ with other TTA proposals. It was observed that the batch size during testing has a significant impact on the direction of stochastic gradient descent, they performed poorly with small batch sizes. In contrast, $\texttt{InPer}$ consistently demonstrated excellent performance. This also highlights the explicit reduction of domain divergence~\cite{zhang2023adanpc} achieved by a parameter-free classifier.

\section{Conclusions}

In this paper, we propose the $\texttt{InPer}$ framework, guided by causal learning, to achieve OOD generalization. The core idea is to decouple domain- and class-related variables using feature representation entropy and to disrupt the relationship between the causal variables and the domain through causal intervention. Furthermore, we unify the training and testing processes and construct a classifier adapted to the target domain using variables stabilized by causal relationships. A limitation that needs to be addressed in future work is the extension of our framework to more backbones, such as vision Transformer and multilayer perceptron. The comprehensive experiments on various benchmarks demonstrate the effectiveness and superiority of our proposal. $\texttt{InPer}$ achieves state-of-the-art performance in a plug-and-play manner, without additional training parameters.

\section*{Acknowledgement}
This work was supported in part by the National Natural Science Foundation of China under Grant 82172033, Grant 52105126, Grant 82272071, and Grant 62271430; and in part by the Dreams Foundation of Jianghuai Advance Technology Center; and in part by the Open Fund of the National Key Laboratory of Infrared Detection Technologies.

\bibliography{egbib}

\begin{thebibliography}{74}
\providecommand{\natexlab}[1]{#1}
\providecommand{\url}[1]{\texttt{#1}}
\expandafter\ifx\csname urlstyle\endcsname\relax
  \providecommand{\doi}[1]{doi: #1}\else
  \providecommand{\doi}{doi: \begingroup \urlstyle{rm}\Url}\fi

\bibitem[Akuzawa et~al.(2020)Akuzawa, Iwasawa, and Matsuo]{akuzawa2020adversarial}
Kei Akuzawa, Yusuke Iwasawa, and Yutaka Matsuo.
\newblock Adversarial invariant feature learning with accuracy constraint for domain generalization.
\newblock In \emph{Machine Learning and Knowledge Discovery in Databases: European Conference, ECML PKDD 2019, W{\"u}rzburg, Germany, September 16--20, 2019, Proceedings, Part II}, pages 315--331, 2020.

\bibitem[Ben-David et~al.(2010)Ben-David, Blitzer, Crammer, Kulesza, Pereira, and Vaughan]{ben2010theory}
Shai Ben-David, John Blitzer, Koby Crammer, Alex Kulesza, Fernando Pereira, and Jennifer~Wortman Vaughan.
\newblock A theory of learning from different domains.
\newblock \emph{Machine learning}, 79:\penalty0 151--175, 2010.

\bibitem[Boudiaf et~al.(2022)Boudiaf, M{\"{u}}ller, Ayed, and Bertinetto]{boudiaf2022parameter}
Malik Boudiaf, Romain M{\"{u}}ller, Ismail~Ben Ayed, and Luca Bertinetto.
\newblock Parameter-free online test-time adaptation.
\newblock In \emph{{IEEE/CVF} Conference on Computer Vision and Pattern Recognition, {CVPR} 2022, New Orleans, LA, USA, June 18-24, 2022}, pages 8334--8343. {IEEE}, 2022.
\newblock \doi{10.1109/CVPR52688.2022.00816}.
\newblock URL \url{https://doi.org/10.1109/CVPR52688.2022.00816}.

\bibitem[Chen et~al.(2022{\natexlab{a}})Chen, Li, Han, Liu, and Yu]{chen2022compound}
Chaoqi Chen, Jiongcheng Li, Xiaoguang Han, Xiaoqing Liu, and Yizhou Yu.
\newblock Compound domain generalization via meta-knowledge encoding.
\newblock In \emph{Proceedings of the IEEE/CVF conference on computer vision and pattern recognition}, pages 7119--7129, 2022{\natexlab{a}}.

\bibitem[Chen et~al.(2022{\natexlab{b}})Chen, Li, Zhou, Han, Huang, Ding, and Yu]{chen2022relation}
Chaoqi Chen, Jiongcheng Li, Hong-Yu Zhou, Xiaoguang Han, Yue Huang, Xinghao Ding, and Yizhou Yu.
\newblock Relation matters: Foreground-aware graph-based relational reasoning for domain adaptive object detection.
\newblock \emph{IEEE Transactions on Pattern Analysis and Machine Intelligence}, 45\penalty0 (3):\penalty0 3677--3694, 2022{\natexlab{b}}.

\bibitem[Chen et~al.(2022{\natexlab{c}})Chen, Tang, Liu, Zhao, Huang, and Yu]{chen2022mix}
Chaoqi Chen, Luyao Tang, Feng Liu, Gangming Zhao, Yue Huang, and Yizhou Yu.
\newblock Mix and reason: Reasoning over semantic topology with data mixing for domain generalization.
\newblock \emph{Advances in Neural Information Processing Systems}, 35:\penalty0 33302--33315, 2022{\natexlab{c}}.

\bibitem[Chen et~al.(2023{\natexlab{a}})Chen, Tang, Huang, Han, and Yu]{chen2023coda}
Chaoqi Chen, Luyao Tang, Yue Huang, Xiaoguang Han, and Yizhou Yu.
\newblock Coda: generalizing to open and unseen domains with compaction and disambiguation.
\newblock \emph{Advances in Neural Information Processing Systems}, 36, 2023{\natexlab{a}}.

\bibitem[Chen et~al.(2023{\natexlab{b}})Chen, Tang, Tao, Zhou, Huang, Han, and Yu]{chen2023activate}
Chaoqi Chen, Luyao Tang, Leitian Tao, Hong-Yu Zhou, Yue Huang, Xiaoguang Han, and Yizhou Yu.
\newblock Activate and reject: towards safe domain generalization under category shift.
\newblock In \emph{Proceedings of the IEEE/CVF International Conference on Computer Vision}, pages 11552--11563, 2023{\natexlab{b}}.

\bibitem[Christiansen et~al.(2020)Christiansen, Pfister, Jakobsen, Gnecco, and Peters]{Christiansen2020ACF}
Rune Christiansen, Niklas Pfister, Martin~Emil Jakobsen, Nicola Gnecco, and J.~Peters.
\newblock A causal framework for distribution generalization.
\newblock \emph{IEEE Transactions on Pattern Analysis and Machine Intelligence}, 44:\penalty0 6614--6630, 2020.

\bibitem[Christiansen et~al.(2021)Christiansen, Pfister, Jakobsen, Gnecco, and Peters]{christiansen2021causal}
Rune Christiansen, Niklas Pfister, Martin~Emil Jakobsen, Nicola Gnecco, and Jonas Peters.
\newblock A causal framework for distribution generalization.
\newblock \emph{IEEE Transactions on Pattern Analysis and Machine Intelligence}, 44\penalty0 (10):\penalty0 6614--6630, 2021.

\bibitem[Cordts et~al.(2016)Cordts, Omran, Ramos, Rehfeld, Enzweiler, Benenson, Franke, Roth, and Schiele]{cordts2016cityscapes}
Marius Cordts, Mohamed Omran, Sebastian Ramos, Timo Rehfeld, Markus Enzweiler, Rodrigo Benenson, Uwe Franke, Stefan Roth, and Bernt Schiele.
\newblock The cityscapes dataset for semantic urban scene understanding.
\newblock In \emph{Proceedings of the IEEE conference on computer vision and pattern recognition}, pages 3213--3223, 2016.

\bibitem[Dou et~al.(2019)Dou, de~Castro, Kamnitsas, and Glocker]{dou2019domain}
Qi~Dou, Daniel~Coelho de~Castro, Konstantinos Kamnitsas, and Ben Glocker.
\newblock Domain generalization via model-agnostic learning of semantic features.
\newblock In Hanna~M. Wallach, Hugo Larochelle, Alina Beygelzimer, Florence d'Alch{\'{e}}{-}Buc, Emily~B. Fox, and Roman Garnett, editors, \emph{Advances in Neural Information Processing Systems 32: Annual Conference on Neural Information Processing Systems 2019, NeurIPS 2019, December 8-14, 2019, Vancouver, BC, Canada}, pages 6447--6458, 2019.
\newblock URL \url{https://proceedings.neurips.cc/paper/2019/hash/2974788b53f73e7950e8aa49f3a306db-Abstract.html}.

\bibitem[Ganin and Lempitsky(2015)]{10.5555/3045118.3045244}
Yaroslav Ganin and Victor Lempitsky.
\newblock Unsupervised domain adaptation by backpropagation.
\newblock In \emph{Proceedings of the 32nd International Conference on International Conference on Machine Learning - Volume 37}, ICML'15, page 1180–1189. JMLR.org, 2015.

\bibitem[Ghifary et~al.(2015)Ghifary, Kleijn, Zhang, and Balduzzi]{Ghifary2015DomainGF}
Muhammad Ghifary, W.~Kleijn, Mengjie Zhang, and David Balduzzi.
\newblock Domain generalization for object recognition with multi-task autoencoders.
\newblock \emph{2015 IEEE International Conference on Computer Vision (ICCV)}, pages 2551--2559, 2015.

\bibitem[Gong et~al.(2016)Gong, Zhang, Liu, Tao, Glymour, and Sch{\"o}lkopf]{gong2016domain}
Mingming Gong, Kun Zhang, Tongliang Liu, Dacheng Tao, Clark Glymour, and Bernhard Sch{\"o}lkopf.
\newblock Domain adaptation with conditional transferable components.
\newblock In \emph{International conference on machine learning}, pages 2839--2848. PMLR, 2016.

\bibitem[Gong et~al.(2018)Gong, Li, Chen, and Gool]{Gong2018DLOWDF}
Rui Gong, Wen Li, Yuhua Chen, and Luc~Van Gool.
\newblock Dlow: Domain flow for adaptation and generalization.
\newblock \emph{2019 IEEE/CVF Conference on Computer Vision and Pattern Recognition (CVPR)}, pages 2472--2481, 2018.

\bibitem[Heinze-Deml and Meinshausen(2021)]{heinze2021conditional}
Christina Heinze-Deml and Nicolai Meinshausen.
\newblock Conditional variance penalties and domain shift robustness.
\newblock \emph{Mach. Learn.}, 110\penalty0 (2):\penalty0 303--348, 2021.
\newblock URL \url{https://doi.org/10.1007/s10994-020-05924-1}.

\bibitem[Hendrycks and Dietterich()]{hendrycks2019benchmarking}
Dan Hendrycks and Thomas~G. Dietterich.
\newblock Benchmarking neural network robustness to common corruptions and perturbations.
\newblock In \emph{7th International Conference on Learning Representations, {ICLR} 2019}.

\bibitem[Huang et~al.(2020)Huang, Wang, Xing, and Huang]{huang2020self}
Zeyi Huang, Haohan Wang, Eric~P Xing, and Dong Huang.
\newblock Self-challenging improves cross-domain generalization.
\newblock In \emph{Computer Vision--ECCV 2020: 16th European Conference, Glasgow, UK, August 23--28, 2020, Proceedings, Part II 16}, pages 124--140. Springer, 2020.

\bibitem[Iwasawa and Matsuo(2021)]{iwasawa2021test}
Yusuke Iwasawa and Yutaka Matsuo.
\newblock Test-time classifier adjustment module for model-agnostic domain generalization.
\newblock In Marc'Aurelio Ranzato, Alina Beygelzimer, Yann~N. Dauphin, Percy Liang, and Jennifer~Wortman Vaughan, editors, \emph{Advances in Neural Information Processing Systems 34: Annual Conference on Neural Information Processing Systems 2021, NeurIPS 2021, December 6-14, 2021, virtual}, pages 2427--2440, 2021.
\newblock URL \url{https://proceedings.neurips.cc/paper/2021/hash/1415fe9fea0fa1e45dddcff5682239a0-Abstract.html}.

\bibitem[Jin et~al.(2020)Jin, Lan, Zeng, Chen, and Zhang]{Jin2020StyleNA}
Xin Jin, Cuiling Lan, Wenjun Zeng, Zhibo Chen, and Li~Zhang.
\newblock Style normalization and restitution for generalizable person re-identification.
\newblock \emph{2020 IEEE/CVF Conference on Computer Vision and Pattern Recognition (CVPR)}, pages 3140--3149, 2020.

\bibitem[Kim et~al.(2021)Kim, Yoo, Park, Kim, and Lee]{kim2021selfreg}
Daehee Kim, Youngjun Yoo, Seunghyun Park, Jinkyu Kim, and Jaekoo Lee.
\newblock Selfreg: Self-supervised contrastive regularization for domain generalization.
\newblock In \emph{Proceedings of the IEEE/CVF International Conference on Computer Vision}, pages 9619--9628, 2021.

\bibitem[Koh et~al.(2020)Koh, Sagawa, Marklund, Xie, Zhang, Balsubramani, Hu, Yasunaga, Phillips, Beery, Leskovec, Kundaje, Pierson, Levine, Finn, and Liang]{Koh2020WILDSAB}
Pang~Wei Koh, Shiori Sagawa, Henrik Marklund, Sang~Michael Xie, Marvin Zhang, Akshay Balsubramani, Weihua Hu, Michihiro Yasunaga, Richard~L. Phillips, Sara Beery, Jure Leskovec, Anshul Kundaje, Emma Pierson, Sergey Levine, Chelsea Finn, and Percy Liang.
\newblock Wilds: A benchmark of in-the-wild distribution shifts.
\newblock In \emph{International Conference on Machine Learning}, 2020.

\bibitem[Komodakis and Gidaris(2018)]{komodakis2018unsupervised}
Nikos Komodakis and Spyros Gidaris.
\newblock Unsupervised representation learning by predicting image rotations.
\newblock In \emph{International conference on learning representations (ICLR)}, 2018.

\bibitem[Kruglanski(1975)]{kruglanski1975endogenous}
Arie~W Kruglanski.
\newblock The endogenous-exogenous partition in attribution theory.
\newblock \emph{Psychological Review}, 82\penalty0 (6):\penalty0 387, 1975.

\bibitem[Li et~al.(2021)Li, Zhang, Liang, Ma, Huang, and Ding]{li2021consistent}
Chenxin Li, Yunlong Zhang, Zhehan Liang, Wenao Ma, Yue Huang, and Xinghao Ding.
\newblock Consistent posterior distributions under vessel-mixing: a regularization for cross-domain retinal artery/vein classification.
\newblock In \emph{2021 IEEE International Conference on Image Processing (ICIP)}, pages 61--65. IEEE, 2021.

\bibitem[Li et~al.(2022{\natexlab{a}})Li, Lin, Mao, Lin, Qi, Ding, Huang, Liang, and Yu]{li2022domain}
Chenxin Li, Xin Lin, Yijin Mao, Wei Lin, Qi~Qi, Xinghao Ding, Yue Huang, Dong Liang, and Yizhou Yu.
\newblock Domain generalization on medical imaging classification using episodic training with task augmentation.
\newblock \emph{Computers in biology and medicine}, 141:\penalty0 105144, 2022{\natexlab{a}}.

\bibitem[Li et~al.(2017)Li, Yang, Song, and Hospedales]{li2017deeper}
Da~Li, Yongxin Yang, Yi-Zhe Song, and Timothy~M Hospedales.
\newblock Deeper, broader and artier domain generalization.
\newblock In \emph{Proceedings of the IEEE international conference on computer vision}, pages 5542--5550, 2017.

\bibitem[Li et~al.(2018)Li, Pan, Wang, and Kot]{li2018domain}
Haoliang Li, Sinno~Jialin Pan, Shiqi Wang, and Alex~C. Kot.
\newblock Domain generalization with adversarial feature learning.
\newblock In \emph{2018 {IEEE} Conference on Computer Vision and Pattern Recognition, {CVPR} 2018, Salt Lake City, UT, USA, June 18-22, 2018}, pages 5400--5409. {IEEE} Computer Society, 2018.
\newblock \doi{10.1109/CVPR.2018.00566}.
\newblock URL \url{http://openaccess.thecvf.com/content\_cvpr\_2018/html/Li\_Domain\_Generalization\_With\_CVPR\_2018\_paper.html}.

\bibitem[Li et~al.(2022{\natexlab{b}})Li, Dai, Ge, Liu, Shan, and Duan]{li2022uncertainty}
Xiaotong Li, Yongxing Dai, Yixiao Ge, Jun Liu, Ying Shan, and Lingyu Duan.
\newblock Uncertainty modeling for out-of-distribution generalization.
\newblock In \emph{The Tenth International Conference on Learning Representations, {ICLR} 2022, Virtual Event, April 25-29, 2022}. ICLR, 2022{\natexlab{b}}.

\bibitem[Liang et~al.(2020)Liang, Hu, and Feng]{liang2020we}
Jian Liang, Dapeng Hu, and Jiashi Feng.
\newblock Do we really need to access the source data? source hypothesis transfer for unsupervised domain adaptation.
\newblock In \emph{International Conference on Machine Learning}, pages 6028--6039. PMLR, 2020.

\bibitem[Liu et~al.(2021{\natexlab{a}})Liu, Hu, Cui, Li, and Shen]{liu2021heterogeneous}
Jiashuo Liu, Zheyuan Hu, Peng Cui, Bo~Li, and Zheyan Shen.
\newblock Heterogeneous risk minimization.
\newblock In Marina Meila and Tong Zhang, editors, \emph{Proceedings of the 38th International Conference on Machine Learning, {ICML} 2021, 18-24 July 2021, Virtual Event}, volume 139 of \emph{Proceedings of Machine Learning Research}, pages 6804--6814. {PMLR}, 2021{\natexlab{a}}.
\newblock URL \url{http://proceedings.mlr.press/v139/liu21h.html}.

\bibitem[Liu et~al.(2022)Liu, Huang, Li, Zheng, and Zha]{DBLP:conf/aaai/Liu0LZZ22}
Jiawei Liu, Zhipeng Huang, Liang Li, Kecheng Zheng, and Zheng{-}Jun Zha.
\newblock Debiased batch normalization via gaussian process for generalizable person re-identification.
\newblock In \emph{{AAAI}}, pages 1729--1737. {AAAI} Press, 2022.

\bibitem[Liu et~al.(2021{\natexlab{b}})Liu, Kothari, van Delft, Bellot{-}Gurlet, Mordan, and Alahi]{liu2021ttt++}
Yuejiang Liu, Parth Kothari, Bastien van Delft, Baptiste Bellot{-}Gurlet, Taylor Mordan, and Alexandre Alahi.
\newblock {TTT++:} when does self-supervised test-time training fail or thrive?
\newblock In Marc'Aurelio Ranzato, Alina Beygelzimer, Yann~N. Dauphin, Percy Liang, and Jennifer~Wortman Vaughan, editors, \emph{Advances in Neural Information Processing Systems 34: Annual Conference on Neural Information Processing Systems 2021, NeurIPS 2021, December 6-14, 2021, virtual}, pages 21808--21820, 2021{\natexlab{b}}.
\newblock URL \url{https://proceedings.neurips.cc/paper/2021/hash/b618c3210e934362ac261db280128c22-Abstract.html}.

\bibitem[Lv et~al.(2022)Lv, Liang, Li, Zang, Liu, Wang, and Liu]{lv2022causality}
Fangrui Lv, Jian Liang, Shuang Li, Bin Zang, Chi~Harold Liu, Ziteng Wang, and Di~Liu.
\newblock Causality inspired representation learning for domain generalization.
\newblock In \emph{Proceedings of the IEEE/CVF Conference on Computer Vision and Pattern Recognition}, pages 8046--8056, 2022.

\bibitem[Magliacane et~al.(2018)Magliacane, Van~Ommen, Claassen, Bongers, Versteeg, and Mooij]{magliacane2018domain}
Sara Magliacane, Thijs Van~Ommen, Tom Claassen, Stephan Bongers, Philip Versteeg, and Joris~M Mooij.
\newblock Domain adaptation by using causal inference to predict invariant conditional distributions.
\newblock \emph{Advances in neural information processing systems}, 31, 2018.

\bibitem[Mahajan et~al.(2021)Mahajan, Tople, and Sharma]{mahajan2021domain}
Divyat Mahajan, Shruti Tople, and Amit Sharma.
\newblock Domain generalization using causal matching.
\newblock In Marina Meila and Tong Zhang, editors, \emph{Proceedings of the 38th International Conference on Machine Learning, {ICML} 2021, 18-24 July 2021, Virtual Event}, volume 139 of \emph{Proceedings of Machine Learning Research}, pages 7313--7324. {PMLR}, 2021.
\newblock URL \url{http://proceedings.mlr.press/v139/mahajan21b.html}.

\bibitem[Mancini et~al.(2018)Mancini, Bul{\`o}, Caputo, and Ricci]{Mancini2018BestSF}
Massimiliano Mancini, Samuel~Rota Bul{\`o}, Barbara Caputo, and Elisa Ricci.
\newblock Best sources forward: Domain generalization through source-specific nets.
\newblock \emph{2018 25th IEEE International Conference on Image Processing (ICIP)}, pages 1353--1357, 2018.

\bibitem[Meng et~al.(2022)Meng, Li, Chen, Yang, Song, Wang, Zhang, Song, Xie, and Pu]{meng2022attention}
Rang Meng, Xianfeng Li, Weijie Chen, Shicai Yang, Jie Song, Xinchao Wang, Lei Zhang, Mingli Song, Di~Xie, and Shiliang Pu.
\newblock Attention diversification for domain generalization.
\newblock In \emph{Computer Vision--ECCV 2022: 17th European Conference, Tel Aviv, Israel, October 23--27, 2022, Proceedings, Part XXXIV}, pages 322--340. Springer, 2022.

\bibitem[Motiian et~al.(2017)Motiian, Piccirilli, Adjeroh, and Doretto]{Motiian2017UnifiedDS}
Saeid Motiian, Marco Piccirilli, Donald~A. Adjeroh, and Gianfranco Doretto.
\newblock Unified deep supervised domain adaptation and generalization.
\newblock \emph{2017 IEEE International Conference on Computer Vision (ICCV)}, pages 5716--5726, 2017.

\bibitem[Nam et~al.(2021)Nam, Lee, Park, Yoon, and Yoo]{nam2021reducing}
Hyeonseob Nam, HyunJae Lee, Jongchan Park, Wonjun Yoon, and Donggeun Yoo.
\newblock Reducing domain gap by reducing style bias.
\newblock In \emph{Proceedings of the IEEE/CVF Conference on Computer Vision and Pattern Recognition}, pages 8690--8699, 2021.

\bibitem[Nazari and Kovashka(2020)]{Nazari2020DomainGU}
Narges~Honarvar Nazari and Adriana Kovashka.
\newblock Domain generalization using shape representation.
\newblock In \emph{ECCV Workshops}, 2020.

\bibitem[Niu et~al.(2022)Niu, Wu, Zhang, Chen, Zheng, Zhao, and Tan]{niu2022efficient}
Shuaicheng Niu, Jiaxiang Wu, Yifan Zhang, Yaofo Chen, Shijian Zheng, Peilin Zhao, and Mingkui Tan.
\newblock Efficient test-time model adaptation without forgetting.
\newblock In Kamalika Chaudhuri, Stefanie Jegelka, Le~Song, Csaba Szepesv{\'{a}}ri, Gang Niu, and Sivan Sabato, editors, \emph{International Conference on Machine Learning, {ICML} 2022, 17-23 July 2022, Baltimore, Maryland, {USA}}, volume 162 of \emph{Proceedings of Machine Learning Research}, pages 16888--16905. {PMLR}, 2022.
\newblock URL \url{https://proceedings.mlr.press/v162/niu22a.html}.

\bibitem[Nuriel et~al.(2021)Nuriel, Benaim, and Wolf]{nuriel2021permuted}
Oren Nuriel, Sagie Benaim, and Lior Wolf.
\newblock Permuted adain: Reducing the bias towards global statistics in image classification.
\newblock In \emph{Proceedings of the IEEE/CVF Conference on Computer Vision and Pattern Recognition}, pages 9482--9491, 2021.

\bibitem[Pan et~al.(2018)Pan, Luo, Shi, and Tang]{DBLP:journals/corr/abs-1807-09441}
Xingang Pan, Ping Luo, Jianping Shi, and Xiaoou Tang.
\newblock Two at once: Enhancing learning and generalization capacities via ibn-net.
\newblock \emph{ECCV}, 2018.

\bibitem[Pearl and Mackenzie(2018)]{pearl2018book}
J.~Pearl and D.~Mackenzie.
\newblock \emph{The Book of Why: The New Science of Cause and Effect}.
\newblock Penguin Books Limited, 2018.
\newblock ISBN 9780241242643.
\newblock URL \url{https://books.google.de/books?id=EmY8DwAAQBAJ}.

\bibitem[Pearl(2009)]{pearl2009causality}
Judea Pearl.
\newblock \emph{Causality}.
\newblock Cambridge university press, 2009.

\bibitem[Peters et~al.(2016)Peters, B{\"u}hlmann, and Meinshausen]{peters2016causal}
Jonas Peters, Peter B{\"u}hlmann, and Nicolai Meinshausen.
\newblock Causal inference by using invariant prediction: identification and confidence intervals.
\newblock \emph{Journal of the Royal Statistical Society. Series B (Statistical Methodology)}, pages 947--1012, 2016.

\bibitem[Peters et~al.(2017)Peters, Janzing, and Sch{\"o}lkopf]{peters2017elements}
Jonas Peters, Dominik Janzing, and Bernhard Sch{\"o}lkopf.
\newblock \emph{Elements of causal inference: foundations and learning algorithms}.
\newblock The MIT Press, 2017.

\bibitem[Recht et~al.(2019)Recht, Roelofs, Schmidt, and Shankar]{recht2019imagenet}
Benjamin Recht, Rebecca Roelofs, Ludwig Schmidt, and Vaishaal Shankar.
\newblock Do imagenet classifiers generalize to imagenet?
\newblock In Kamalika Chaudhuri and Ruslan Salakhutdinov, editors, \emph{Proceedings of the 36th International Conference on Machine Learning, {ICML} 2019, 9-15 June 2019, Long Beach, California, {USA}}, volume~97 of \emph{Proceedings of Machine Learning Research}, pages 5389--5400. {PMLR}, 2019.
\newblock URL \url{http://proceedings.mlr.press/v97/recht19a.html}.

\bibitem[Richter et~al.(2016)Richter, Vineet, Roth, and Koltun]{richter2016playing}
Stephan~R Richter, Vibhav Vineet, Stefan Roth, and Vladlen Koltun.
\newblock Playing for data: Ground truth from computer games.
\newblock In \emph{Computer Vision--ECCV 2016: 14th European Conference, Amsterdam, The Netherlands, October 11-14, 2016, Proceedings, Part II 14}, pages 102--118. Springer, 2016.

\bibitem[Rojas-Carulla et~al.(2018)Rojas-Carulla, Sch{\"o}lkopf, Turner, and Peters]{rojas2018invariant}
Mateo Rojas-Carulla, Bernhard Sch{\"o}lkopf, Richard Turner, and Jonas Peters.
\newblock Invariant models for causal transfer learning.
\newblock \emph{The Journal of Machine Learning Research}, 19\penalty0 (1):\penalty0 1309--1342, 2018.

\bibitem[Sch\"{o}lkopf(2022)]{10.1145/3501714.3501755}
Bernhard Sch\"{o}lkopf.
\newblock \emph{Causality for Machine Learning}, page 765–804.
\newblock Association for Computing Machinery, New York, NY, USA, 1 edition, 2022.
\newblock ISBN 9781450395861.
\newblock URL \url{https://doi.org/10.1145/3501714.3501755}.

\bibitem[Seo et~al.(2020)Seo, Suh, Kim, Han, and Han]{DBLP:journals/corr/abs-1907-04275}
Seonguk Seo, Yumin Suh, Dongwan Kim, Jongwoo Han, and Bohyung Han.
\newblock Learning to optimize domain specific normalization for domain generalization.
\newblock \emph{ECCV}, page 68–83, 2020.

\bibitem[Shankar et~al.(2018)Shankar, Piratla, Chakrabarti, Chaudhuri, Jyothi, and Sarawagi]{Shankar2018GeneralizingAD}
Shiv Shankar, Vihari Piratla, Soumen Chakrabarti, Siddhartha Chaudhuri, Preethi Jyothi, and Sunita Sarawagi.
\newblock Generalizing across domains via cross-gradient training.
\newblock \emph{ICLR}, 2018.

\bibitem[Shannon(2001)]{shannon2001mathematical}
Claude~Elwood Shannon.
\newblock A mathematical theory of communication.
\newblock \emph{ACM SIGMOBILE mobile computing and communications review}, 5\penalty0 (1):\penalty0 3--55, 2001.

\bibitem[Subbaswamy et~al.(2019)Subbaswamy, Schulam, and Saria]{subbaswamy2019preventing}
Adarsh Subbaswamy, Peter Schulam, and Suchi Saria.
\newblock Preventing failures due to dataset shift: Learning predictive models that transport.
\newblock In Kamalika Chaudhuri and Masashi Sugiyama, editors, \emph{The 22nd International Conference on Artificial Intelligence and Statistics, {AISTATS} 2019, 16-18 April 2019, Naha, Okinawa, Japan}, volume~89 of \emph{Proceedings of Machine Learning Research}, pages 3118--3127. {PMLR}, 2019.
\newblock URL \url{http://proceedings.mlr.press/v89/subbaswamy19a.html}.

\bibitem[Sun et~al.(2020)Sun, Wang, Liu, Miller, Efros, and Hardt]{sun2020test}
Yu~Sun, Xiaolong Wang, Zhuang Liu, John Miller, Alexei~A. Efros, and Moritz Hardt.
\newblock Test-time training with self-supervision for generalization under distribution shifts.
\newblock In \emph{Proceedings of the 37th International Conference on Machine Learning, {ICML} 2020, 13-18 July 2020, Virtual Event}, volume 119 of \emph{Proceedings of Machine Learning Research}, pages 9229--9248. {PMLR}, 2020.
\newblock URL \url{http://proceedings.mlr.press/v119/sun20b.html}.

\bibitem[Tremblay et~al.(2018)Tremblay, Prakash, Acuna, Brophy, Jampani, Anil, To, Cameracci, Boochoon, and Birchfield]{Tremblay2018TrainingDN}
Jonathan Tremblay, Aayush Prakash, David Acuna, Mark Brophy, V.~Jampani, Cem Anil, Thang To, Eric Cameracci, Shaad Boochoon, and Stan Birchfield.
\newblock Training deep networks with synthetic data: Bridging the reality gap by domain randomization.
\newblock \emph{2018 IEEE/CVF Conference on Computer Vision and Pattern Recognition Workshops (CVPRW)}, pages 1082--10828, 2018.

\bibitem[Van~der Maaten and Hinton(2008)]{van2008visualizing}
Laurens Van~der Maaten and Geoffrey Hinton.
\newblock Visualizing data using t-sne.
\newblock \emph{Journal of machine learning research}, 9\penalty0 (11), 2008.

\bibitem[Vapnik(1991)]{vapnik1991principles}
Vladimir Vapnik.
\newblock Principles of risk minimization for learning theory.
\newblock \emph{Advances in neural information processing systems}, 4, 1991.

\bibitem[Venkateswara et~al.(2017)Venkateswara, Eusebio, Chakraborty, and Panchanathan]{venkateswara2017deep}
Hemanth Venkateswara, Jose Eusebio, Shayok Chakraborty, and Sethuraman Panchanathan.
\newblock Deep hashing network for unsupervised domain adaptation.
\newblock In \emph{Proceedings of the IEEE conference on computer vision and pattern recognition}, pages 5018--5027, 2017.

\bibitem[Wang et~al.(2021)Wang, Shelhamer, Liu, Olshausen, and Darrell]{wang2020tent}
Dequan Wang, Evan Shelhamer, Shaoteng Liu, Bruno~A. Olshausen, and Trevor Darrell.
\newblock Tent: Fully test-time adaptation by entropy minimization.
\newblock ICLR, 2021.

\bibitem[Wang et~al.(2022{\natexlab{a}})Wang, Fink, Van~Gool, and Dai]{wang2022continual}
Qin Wang, Olga Fink, Luc Van~Gool, and Dengxin Dai.
\newblock Continual test-time domain adaptation.
\newblock In \emph{Proceedings of the IEEE/CVF Conference on Computer Vision and Pattern Recognition}, pages 7201--7211, 2022{\natexlab{a}}.

\bibitem[Wang et~al.(2022{\natexlab{b}})Wang, Liu, Chen, Wu, Hao, Chen, and Heng]{wang2022contrastive}
Yunqi Wang, Furui Liu, Zhitang Chen, Yik-Chung Wu, Jianye Hao, Guangyong Chen, and Pheng-Ann Heng.
\newblock Contrastive-ace: Domain generalization through alignment of causal mechanisms.
\newblock \emph{IEEE Transactions on Image Processing}, 32:\penalty0 235--250, 2022{\natexlab{b}}.

\bibitem[Wei et~al.(2021)Wei, Lan, Zeng, and Chen]{Wei2021ToAlignTA}
Guoqiang Wei, Cuiling Lan, Wenjun Zeng, and Zhibo Chen.
\newblock Toalign: Task-oriented alignment for unsupervised domain adaptation.
\newblock In \emph{Neural Information Processing Systems}, 2021.

\bibitem[Yue et~al.(2019)Yue, Zhang, Zhao, Sangiovanni-Vincentelli, Keutzer, and Gong]{Yue2019DomainRA}
Xiangyu Yue, Yang Zhang, Sicheng Zhao, Alberto~L. Sangiovanni-Vincentelli, Kurt Keutzer, and Boqing Gong.
\newblock Domain randomization and pyramid consistency: Simulation-to-real generalization without accessing target domain data.
\newblock \emph{2019 IEEE/CVF International Conference on Computer Vision (ICCV)}, pages 2100--2110, 2019.

\bibitem[Zhang et~al.(2018)Zhang, Cisse, Dauphin, and Lopez-Paz]{zhang2018mixup}
Hongyi Zhang, Moustapha Cisse, Yann~N. Dauphin, and David Lopez-Paz.
\newblock mixup: Beyond empirical risk minimization.
\newblock In \emph{International Conference on Learning Representations}. ICLR, 2018.

\bibitem[Zhang et~al.(2022)Zhang, Li, Li, Jia, and Zhang]{zhang2022exact}
Yabin Zhang, Minghan Li, Ruihuang Li, Kui Jia, and Lei Zhang.
\newblock Exact feature distribution matching for arbitrary style transfer and domain generalization.
\newblock In \emph{Proceedings of the IEEE/CVF Conference on Computer Vision and Pattern Recognition}, pages 8035--8045, 2022.

\bibitem[Zhang et~al.(2023)Zhang, Wang, Jin, Yuan, Zhang, Wang, Jin, and Tan]{zhang2023adanpc}
Yi-Fan Zhang, Xue Wang, Kexin Jin, Kun Yuan, Zhang Zhang, Liang Wang, Rong Jin, and Tieniu Tan.
\newblock Adanpc: Exploring non-parametric classifier for test-time adaptation.
\newblock \emph{arXiv preprint arXiv:2304.12566}, 2023.

\bibitem[Zhou et~al.(2020{\natexlab{a}})Zhou, Yang, Hospedales, and Xiang]{zhou2020learning}
Kaiyang Zhou, Yongxin Yang, Timothy Hospedales, and Tao Xiang.
\newblock Learning to generate novel domains for domain generalization.
\newblock In \emph{Computer Vision--ECCV 2020: 16th European Conference, Glasgow, UK, August 23--28, 2020, Proceedings, Part XVI 16}, pages 561--578. ECCV, 2020{\natexlab{a}}.

\bibitem[Zhou et~al.(2020{\natexlab{b}})Zhou, Yang, Hospedales, and Xiang]{Zhou2020DeepDI}
Kaiyang Zhou, Yongxin Yang, Timothy~M. Hospedales, and Tao Xiang.
\newblock Deep domain-adversarial image generation for domain generalisation.
\newblock \emph{AAAI}, page 13025–13032, 2020{\natexlab{b}}.

\bibitem[Zhou et~al.(2020{\natexlab{c}})Zhou, Yang, Qiao, and Xiang]{Zhou2020DomainAE}
Kaiyang Zhou, Yongxin Yang, Yu~Qiao, and Tao Xiang.
\newblock Domain adaptive ensemble learning.
\newblock \emph{IEEE Transactions on Image Processing}, 30:\penalty0 8008--8018, 2020{\natexlab{c}}.

\bibitem[Zhou et~al.(2021)Zhou, Yang, Qiao, and Xiang]{zhou2021domain}
Kaiyang Zhou, Yongxin Yang, Yu~Qiao, and Tao Xiang.
\newblock Domain generalization with mixstyle.
\newblock In \emph{9th International Conference on Learning Representations, {ICLR} 2021, Virtual Event, Austria, May 3-7, 2021}. ICLR, 2021.

\end{thebibliography}
\end{document}